\documentclass{article}
\usepackage{ifthen}
\newboolean{preprint}
\setboolean{preprint}{true}

\ifthenelse{\boolean{preprint}}{\usepackage[preprint]{neurips_2021}}{\usepackage{neurips_2021}}

\usepackage[utf8]{inputenc}
\usepackage{amsmath, amsthm, amsfonts, graphicx, amssymb, tabularx}
\usepackage[colorlinks=true,linkcolor=MidnightBlue,urlcolor=MidnightBlue,citecolor=MidnightBlue]{hyperref}

\usepackage{multirow,bigdelim,bm,dsfont,mathtools}
\usepackage{adjustbox}
\usepackage[dvipsnames]{xcolor}
\usepackage{pgfplots, xspace, float, subcaption}
\usepackage[title]{appendix}
\usepackage{etoolbox, booktabs, multirow}
\usepackage{wrapfig}
\usepackage[capitalise,noabbrev]{cleveref}
\usepackage{algorithm}
\usepackage[noend]{algpseudocode}
\usepackage[utf8]{inputenc}

\DeclareUnicodeCharacter{2212}{−}
\usepgfplotslibrary{groupplots,dateplot}
\usetikzlibrary{patterns,shapes.arrows}
\pgfplotsset{compat=newest}

\LetLtxMacro{\ORIcitep}{\citep}
\DeclareRobustCommand{\citep}{\leavevmode\unskip~\ORIcitep}

\LetLtxMacro{\ORIcitet}{\citet}
\DeclareRobustCommand{\citet}{\leavevmode\unskip~\ORIcitet}

\newcommand{\source}{\ensuremath{\mathcal{S}}\xspace}
\newcommand{\target}{\ensuremath{\mathcal{T}}\xspace}
\newcommand{\loss}[1]{\ensuremath{\mathcal{L}}(#1)\xspace}

\newcolumntype{Y}{>{\centering\arraybackslash}X}

\newcommand{\xvec}{\ensuremath{\bm{x}}\xspace}
\newcommand{\xtilde}{\ensuremath{\widetilde{\xvec}}\xspace}
\DeclarePairedDelimiterX{\infdivx}[2]{(}{)}{%
  #1\;\delimsize\|\;#2%
}

\newcommand{\kld}[2]{\ensuremath{D_{KL}\infdivx{#1}{#2}}\xspace}

\newcolumntype{Z}{>{\setbox0=\hbox\bgroup}c<{\egroup}@{\hspace*{-\tabcolsep}}}

\definecolor{color1}{rgb}{0.00392156862745098, 0.45098039215686275, 0.6980392156862745}
\definecolor{color2}{rgb}{0.8705882352941177, 0.5607843137254902, 0.0196078431372549}
\definecolor{color3}{rgb}{0.00784313725490196, 0.6196078431372549, 0.45098039215686275}
\definecolor{color4}{rgb}{0.8352941176470589, 0.3686274509803922, 0.0}
\title{Test time Adaptation through Perturbation Robustness}
\author{Prabhu Teja S\\
Idiap Research Institute \& EPFL, \\
Switzerland\\
\texttt{prabhu.teja@idiap.ch}
\And Fran\c{c}ois Fleuret \\
Idiap Research Institute \& University of Geneva, \\
Switzerland\\
\texttt{francois.fleuret@unige.ch}}
\date{}

\begin{document}
\maketitle

\begin{abstract}
    Data samples generated by several real world processes are dynamic in nature i.e., their characteristics vary with time.
    Thus it is not possible to train and tackle all possible distributional shifts between training and inference, using the
    host of transfer learning methods in literature. In this paper, we tackle this problem of adapting to domain shift at
    inference time i.e., we do not change the training process, but quickly adapt the model at test-time to handle any
    domain shift. For this, we propose to enforce consistency of predictions of data sampled in the vicinity of test sample
    on the image manifold. On a host of test scenarios like dealing with corruptions (CIFAR-10-C and CIFAR-100-C), and
    domain adaptation (VisDA-C), our method is at par or significantly outperforms previous methods.
\end{abstract}

\section{Introduction}
One of the implicit assumptions in the development of machine learning systems is that test data and train data are
sampled from the same distribution. A minor departure from this assumption can have catastrophic
consequences\citep{recht2019imagenet,hendrycks2019robustness,jo2017measuring}. Significant amount of literature has
concentrated on tackling this problem in several ways. Methods in \textit{transfer learning} modify a source trained
network using labeled target domain training data so that performance on test data, which is sampled from the target
domain, is improved\citep{zhuang2021survey}. In problems like semantic segmentation, labeled data is harder to obtain.
For such problems, \textit{Unsupervised domain adaptation} methods have been proposed\citep{toldo2020unsupervised}, in
which unlabeled target data and labeled source domain data are available at train time. Instead of source domain
training data, a source trained model is used in \textit{model
    adaptation}\citep{chidlovskii2016domain,li2020modeladaptation}. A summary of the described scenarios is given in
\cref{tab:problemtypes}. The most oft studied scenarios, like the ones described above, require prior knowledge of the
domain shift that will occur at test time, and require data from the target domain at the train time. For some problems
in sensor data\citep{vegara2020chemical}, autonomous driving\citep{bobu2018adapting}, the domain changes gradually, and
is linked to a physical process (say change of weather, or time of day). In addition to the gradualness, these changes
cannot be always anticipated and thus it is not possible to curate data and train networks for them. This necessitates
training strategies which do not need test domain data at train time.

In this paper, we tackle the problem of handling domain shift at test time; a problem previously termed test time
training\citep{sun2020ttt}, or test time adaptation\citep{wang2021tent}. Differing from model adaptation, the model is
adjusted to (possibly) different domains as the data arrives during inference. Owing to the nature of the problem,
test time adaptation necessitates simpler, lighter methods compared to its train-time adaptation counterparts.

\begin{table}
    \begin{tabularx}{\textwidth}{@{}YccYc@{}}
        \toprule
        Strategy                       & Train data                             & Test data   & Training process                    & Testing  process \\
        \midrule
        Baseline                       & \source                                & $\target_x$ & \loss{\source}                      & -                \\
        Transfer learning              & \source \& \target                     & $\target_x$ & \loss{\source} + \loss{\target}     & -                \\
        Fine tuning                    & \target \& $f(\cdot, \theta_S)$        & $\target_x$ & \loss{\target}                      & -                \\
        Unsupervised Domain Adaptation & \source \& $\target_{x'}$              & $\target_x$ & \loss{\source} + $\loss{\target_x}$ & -                \\
        Model Adaptation               & $\target_{x'}$ \& $f(\cdot, \theta_S)$ & $\target_x$ & $\loss{\target_x}$                  & -                \\
        \textbf{Test time adaptation}  & \source                                & $\target_x$ & \loss{\source}                      & \loss{\target}   \\
        \midrule
    \end{tabularx}
    \caption{In the table we show the various scenarios proposed to handle domain change problem. $\mathcal{X}_S$,
    $\mathcal{X}_T$ denote the source and target domains, and $\source = \{(x_i, y_i)\}_{i=1}^{N_s}$ $\target = \{(x_i,
        y_i)\}_{i=1}^{N_t}$ be the labeled source and target data, and $\source_x$ and $\target_x/\target_{x'}$ are the
    unlabeled data from source and target domains, $\loss{\cdot}$ denotes a generic applicable loss function. Testing
    process with `-' denotes that the testing procedure does not have steps other than the forward-propagation of the data
    sample. In this work, we focus on the last row i.e, test time adaptation, where we push the burden of handling domain
    changes to inference time. A similar table has been presented by \citet{wang2021tent}}
    \label{tab:problemtypes}
\end{table}

Several works have shown that one of ways to improve generalization performance is to train the network with various
input data augmentations\citep{hendrycks2020augmix,cubuk2019randaugment}. Networks are trained with multiple such
augmentations for each training sample with the original label. This is equivalent to training the network  with
original images, and adding a consistency regularizer that penalizes different outputs for
augmentations\citep{leen1994data}. We take a similar view point, and show that enforcing consistency of predictions over
various augmentations of the test samples improves model's performance on corrupted data
(CIFAR-10-C/100-C\citep{hendrycks2019robustness}), and on domain shifts (VisDA-C\citep{peng2018visda}).

\section{Related work}\label{sec:related_work} 
\paragraph{Data augmentations}
Successes of deep learning can be partly attributed to training recipes like data
augmentation\citep{krizhevsky2012imagenet}, among others. There have been several attempts at designing these
augmentation strategies in the literature. \citet{ratner2017learningtocompose,cubuk2019autoaugment} propose to learn
data augmentation per task using reinforcement learning. Recently, randomized augmentation strategies like
RandAugment\citep{cubuk2019randaugment} and AugMix\citep{hendrycks2020augmix} have been shown to improve generalization
performance, as well as calibration. %
Data augmentations are the cornerstone of the modern self-supervised learning
methods (\Citet{chen2020simple}, and survey blog post\citep{weng2021contrastive}). A comprehensive survey has been
presented in \citet{Shorten2019ASO}.

\paragraph{Consistency losses}
Consistency losses were found to help semi-supervised learning
problems \citep{rasmus2015ladder,tarvainen2017mean,bachmann2014pseudoensembles,sajjadi2016regularization}. The outputs of
the augmented data have been compared through JS-divergence in the current work and in \citet{hendrycks2020augmix},
mean-squared error in \citet{sajjadi2016regularization,laine2017temporal}, cross-entropy loss\citep{miyato2018virtual}.
Consistency losses over augmented inputs have been the mainstay in self-supervised learning
literature like InfoNCE loss\citep{chen2020simple} and cross-correlation\citep{zbontar2021barlow}.

\paragraph{Transfer learning} 
In addition to standard techniques like adversarial adaptation\citep{ganin2016domain}, data augmentation based
techniques have also been proposed to deal with domain transfer. \citet{volpi2018adversarial} propose that adversarial
augmentation leads to better generalization for image domains that are similar to source domain. Pretraining with
auxiliary tasks and using consistency losses was found useful for domain adaptation tasks\citep{mishra2021surprisingly}.
\citet{Huang_2018_ECCV} use an image translation network for data augmentation for domain adaptation. Augmented inputs
have been used to predict multiple outputs, and fused by uncertainty weighting in \citet{yeo2021robustness}.
The problem closest to ours is model adaptation\citep{chidlovskii2016domain}, where several recent works have shown the
utility of pseudo-labeling, and entropy reduction on the unlabeled target set\citep{teja2020uncertainty,liang2020shot}.
In addition, methods that use GANs for image synthesis of target data also exist\citep{li2020modeladaptation}.

\paragraph{Test time adaptation}
\citet{sun2020ttt} proposed to adapt networks at test time by training networks for the main task and a pretext task
like rotation prediction at train time. At test time, they fine-tune the pretext task network to adapt to distribution
changes on-the-fly. This method modifies the training procedure to be able to adapt on-the-fly. Recent
works~\citep{nado2021evaluating,schneider2020betterinc} found that adapting batch normalization statistics to the
test-data is a good baseline for dealing with corruptions at test time. {Tent}\citep{wang2021tent} takes this a step
further and fine-tunes batch normalization's affine parameters with an entropy penalty. Important questions about the
sufficiency of normalization statistics are left unanswered in these works. %
Our work differs from the existing works in test time adaptation, as we propose increasing robustness of the network to
test samples. It has been shown in \citep{novak2018sensitivity} that trained neural networks are robust to input
perturbations sampled in the vicinity of the train data; we extend this notion to test time adatation, by enforcing this
robustness as a proxy for improving performance on the test data.

\section{Proposed method}\label{sec:proposed}

Let $p_\cdot\equiv p(y|\cdot) = f(\cdot, \theta)$ be the trained model parameterized by $\theta$ on the source training
set \textit{i.e.,} $f(\cdot, \theta)$ subsumes the softmax layer. Let \xvec be a test sample and $\xtilde\sim
    \mu(\xtilde | \xvec)$ be an output of data augmentation method $\mu(\cdot|\cdot)$with input $\bm{x}$. For a test sample
\xvec, we sample two augmentations $\xtilde_1, \xtilde_2 \sim \mu(\xtilde | \xvec)$, compute the output probabilities as
$p_{\xvec} = f(\xvec, \theta)$, $p_{\xtilde_1}= f(\xtilde_1, \theta)$, $p_{\xtilde_2}= f(\xtilde_2, \theta)$ for the
original input, and the two augmentations. Over these, we propose to use a consistency loss as in \cref{eqn:consloss}
\begin{equation}\label{eqn:consloss}
    \mathcal{L}_{cons}(p_{\xvec}, p_{\xtilde_1}, p_{\xtilde_2}) = \frac{\kld{p_{\xvec}}{\bar{p}} + \kld{p_{\xtilde_1}}{\bar{p}} + \kld{p_{\xtilde_2}}{\bar{p}}}{3},
\end{equation}
where
\begin{equation}
    \bar{p} = \frac{p_{\xvec} + p_{\xtilde_1} + p_{\xtilde_2}}{3}
\end{equation}
is the average posterior density of predictions. Here \[\kld{p}{q} = \sum _{k}p^k\log \left({\frac {p^k}{q^k}}\right)\]
denotes the KL-divergence of the two distributions ${p}$ and ${q}$, where $p^k$ denotes the index $k$. We note that this
loss was originally proposed in \citet{hendrycks2020augmix} for training with augmentations, and refer the readers to it
for experiments about the specific form of \cref{eqn:consloss}. Mean-squared error over the posterior
predictions\citep{tarvainen2017mean,berthelot2019mixmatch} has also been successful in semi-supervised learning. The
details of augmentation methods ($\mu(\cdot|\cdot)$) used are presented in \cref{app:sec:augmentation}.

In addition to the consistency loss in \cref{eqn:consloss}, we use an entropy penalty
\citep{wang2021tent,vu2018advent,grandvalet2005semi}. Contrary to \citep{wang2021tent}, we tune all the parameters of
the networks considered.
\begin{equation}\label{eqn:entropyloss}
    \mathcal{L}_{ent}(p_{\xvec}) = -\sum_{k} p_{\xvec}^k \log(p_{\xvec}^k)
\end{equation}\vspace{-1pt}
where the index $k$ is over the number of classes. The overall loss is given by the sum of \cref{eqn:consloss} and
\cref{eqn:entropyloss}
\begin{equation}\label{eqn:total}
    \mathcal{L} = \mathcal{L}_{cons} + \mathcal{L}_{ent}
\end{equation}

\section{Experiments}
We show the efficacy of our method on the standard benchmarks used in Tent\citep{wang2021tent} for ease of comparison:
corrupted CIFAR-10 and CIFAR-100, named CIFAR-10-C and CIFAR-100-C respectively\citep{hendrycks2019robustness}, and
VisDA-C\citep{peng2018visda}. CIFAR-10-C and CIFAR-100-C consist 15 corruption types that have been algorithmically
generated to benchmark robustness algorithms. VisDA-C dataset is a large-scale dataset that provides training and
testing data with 12 classes (real world objects). Training data consists of rendered 3D models using varying view-point
and lighting conditions, and the test data is real images cropped from Youtube videos. 

For each experiment, we sample two augmentations using either RandAugment\citep{cubuk2019randaugment} or
AugMix\citep{hendrycks2020augmix}, and minimize the loss in \cref{eqn:total} using SGD optimizer with a learning rate of
$10^{-4}$, momentum $0.9$, and weight decay of $5\times10^{-4}$ for $5$ iterations. Additional details of the
augmentation methods, and datasets are presented in \cref{app:sec:augmentation} and \cref{app:sec:datasets}
respectively. We show our main results in this section, and ablations in \cref{app:sec:ablations}.
\begin{table}[tbp]
    \centering
    \caption{Results of VisDA adaptation. We improve significantly compared to the baseline.}    \label{tab:res:visda}
    \medskip
    \begin{tabularx}{9.5cm}{@{}cccc@{}}
        \toprule
        Unadapted & Tent                 & Proposed      & Proposed \\
                  & \citep{wang2021tent} & (RandAugment) & (AugMix) \\
        \midrule
        $44.1$    & $60.9$               & $67.1$        & $67.2$   \\
        \bottomrule
    \end{tabularx}
\end{table}
\paragraph{Domain Adaptation}
We use a ResNet-50\citep{he2016resnet} network that has been pretrained on Imagenet, and use a test time batch size of
$64$. With the results presented in \cref{tab:res:visda}, we see the proposed method beats the existing methods by a
significant margin of $\sim 6.2\%$. We find that lower intensity of augmentations \textit{i.e.}, $m = 1$, $n=1$ for
Randaugment, and $\alpha=1$, $\mathrm{depth}=3$, $\mathrm{severity}=2$, $\mathrm{width}=1$ for AugMix result in the best
performance. We also found that our method is relatively stable to small changes to augmentation parameters used.

\paragraph{Corruption (CIFAR-10-C and CIFAR-100-C)}
We use Wide ResNet\citep{Zagoruyko2016wrn} pretrained models from the RobustBench code
repository\citep{croce2020robustbench}. For CIFAR-10-C we use a WRN-28-10, and for CIFAR-100-C we use WRN-40-2. These
networks have been trained on training set of CIFAR 10/100. We use the augmentation hyperparameters as in domain
adaptation experiments, with the batch size changed to 200. In \cref{tab:res:cifar} and in \cref{app:sec:cifar}, we see
that the performance of our proposed method is comparable to Tent on CIFAR-10-C, and beats Tent by a margin of $\sim
    1\%$ for CIFAR-100-C.

\begin{table}[H]
    \centering
    \caption{Results of CIFAR-10-C and CIFAR-100-C datasets. We present the \% accuracy averaged over all corruptions here.}
    \label{tab:res:cifar}
    \medskip
    \begin{tabularx}{0.85\textwidth}{@{}lcccc}
        \toprule
        Dataset     & Unadapted & Tent                 & Proposed      & Proposed \\
                    &           & \citep{wang2021tent} & (RandAugment) & (AugMix) \\
        \midrule
        CIFAR-10-C  & $56.5 $   & $81.4$               & $80.9$        & $81.2$   \\
        CIFAR-100-C & $53.2 $   & $64.5$               & $65.4$        & $65.7$   \\
        \bottomrule
    \end{tabularx}
\end{table}

\section{Discussion}
Our experiments show that a consistency regularization on the input space (\cref{eqn:consloss}) improves performance on
the test data. Previous studies found that networks with flat minima in the weight space generalized
better\citep{chaudhari2017entropySGD,keskar2016large}. \citet{huang2020understanding} consider neural networks with flat
minima analogous to wide margin classifiers. Wide margin, in addition to being interpreted as stability to perturbation
of parameters, and can also be seen as being robust to input
perturbations\citep{bousquet2002stability,elsayed2018largemargin}. While these prior works partially explain our
proposed method, they do not explain how solely searching for a flat region (in the input space) corresponds to a point
in the weight space that generalizes. We leave this for future research. The specifics of required augmentation need
further study; we show that our proposed method performs quite well with both the augmentation techniques used, but
questions about what constitutes essential characteristics of augmentations remain. We find that the efficacy of our
method (as well as of Tent\citep{wang2021tent}) is dependent on the batch size used (\cref{app:sec:ablations}), which is
a departure from any normal testing scenario, where each sample is labeled independently. Additionally, our
experimentation is currently limited to smaller datasets, and a limited set of problems. Thus, further experiments on
perturbations, and more difficult domain adaptation problems are left as future work.

\ifthenelse{\boolean{preprint}}{\section*{Acknowledgments}
    The research leading to these results was supported by the ``Swiss
    Center for Drones and Robotics - SCDR'' of the Department of Defence,
    Civil Protection and Sport via armasuisse S+T under project n\textsuperscript{o}050-38.}{}

{\small
    \bibliographystyle{plainnat}
    \bibliography{mybib}
}

\appendix

\newcommand{\augmix}{\textsc{AugMix}\xspace}
\section{Augmentation methods}\label{app:sec:augmentation}
In this section we describe the augmentation methods RandAugment\citep{cubuk2019randaugment}, and
\augmix\citep{hendrycks2020augmix} used widely in the main text.

\subsection{RandAugment}
Methods like AutoAugment\citep{cubuk2019autoaugment} use complex machinery like reinforcement learning to deduce the
optimal augmentation strategy for a problem. RandAugment proposes to do the opposite: it randomly samples augmentations
from a predefined list of augmentations and composes them functionally to output the augmentation for each data point.
The algorithm is shown in \cref{alg:randaugment}.
\begin{algorithm}[H]
    \caption{RandAugment}\label{alg:randaugment}
    \begin{algorithmic}
        \State {\textbf{Input:} $\mathcal{O}$ $\leftarrow$ \{Identity, AutoContrast, Equalize,
            Rotate, Solarize, Color, Posterize,
            Contrast, Brightness, Sharpness,
            ShearX, ShearY, TranslateX, TranslateY\}, \xvec, $m$, $n$\\}\Comment{$m$ is the maximum intensity of augmentations, $n$ is the number
            of augmentations}
        \For{i in 1 \dots $n$}
        \State local\_intensity $\gets$ randint(1, $m$)
        \State sample $\gets$ random\_choice($\mathcal{O}$)
        \State $\xvec\gets$ sample(local\_intensity)(\xvec)
        \EndFor
        \State {\bfseries Return:}  \xvec
    \end{algorithmic}
\end{algorithm}

\subsection{\augmix}
\augmix is a data augmentation technique that has been improve model robustness. Unlike RandAugment, where sampled
augmentations are composed, AugMix mixes the results of chains of augmentations in convex combinations. Increasing
diversity  by composing augmentations can generate a sample that is off the data manifold, and the authors argue that
their proposed way of combining generates realistic transformations. The specific algorithm is given in
\cref{alg:augmix}.
\begin{algorithm}[H]
    \caption{\augmix. Adapted from \citet{hendrycks2020augmix}}\label{alg:augmix}
    \begin{algorithmic}
        \State {\textbf{Input:} $\mathcal{O}$ similar to \cref{alg:randaugment}, $k$, $\alpha$}
        \State $\xvec_{aug} \gets$ zeros\_like(\xvec)
        \State $(w_1, w_2, \dots w_k) \sim$ Dirichlet$(\alpha, \alpha, \dots, \alpha)$
        \For{i in 1 \dots $k$}
        \State sample\textsubscript{1}, sample\textsubscript{2}, sample\textsubscript{3} $\gets$ random\_choice($\mathcal{O}$)
        \State sample\textsubscript{1, 2} $\gets$ sample\textsubscript{1}$\circ$\,sample\textsubscript{2}, sample\textsubscript{1, 2, 3} $\gets$ sample\textsubscript{1}$\circ$\,sample\textsubscript{2}$\circ$\,sample\textsubscript{3}
        \State op $\gets$ random\_choice(\{sample\textsubscript{1}, sample\textsubscript{1, 2}, sample\textsubscript{1, 2, 3}\})
        \State $\xvec_{aug} += w_i\cdot op(\xvec)$
        \EndFor
        \State sample $m\sim\mathrm{Beta}(\alpha, \alpha)$
        \State $\xvec_{\mathrm{augmix}} = m\xvec + (1 - m)\xvec_{aug}$
        \State {\bfseries Return:}  $\xvec_{\mathrm{augmix}}$
    \end{algorithmic}
\end{algorithm}

\section{Datasets}\label{app:sec:datasets}
\subsection{Corruption datasets -- CIFAR-10-C, CIFAR-100-C}
To benchmark the model robustness \citet{hendrycks2019robustness} proposed corruption (along with perturbation)
datasets. In this work we consider the datasets that were derived from the commonly used CIFAR-10 and CIFAR-100
datasets, named CIFAR-10C and CIFAR100-C respectively. They consist 15 corruptions types applied to the test data of the
original datasets. Each of the 15 corruptions have 5 levels of severity, and we present our results on the maximum level
of corruption. The corruptions can be grouped into four main categories -- noise, blur, weather and digital. A summary
is presented in \cref{app:tab:corruptions}.
\begin{table}[H]
    \centering
    \caption{Summary of corruptions in CIFAR-10-C and CIFAR-100-C used in this work.}\label{app:tab:corruptions} %
    \begin{tabular}{cccc}
        \toprule
        \textbf{Noise} & \textbf{Blur}      & \textbf{Weather} & \textbf{Digital}       \\
        \midrule
        Gaussian noise & Defocus blur       & Snow             & Brightness             \\
        Shot noise     & Frosted Glass blur & Frost            & Contrast               \\
        Impulse noise  & Motion blur        & Fog              & Elastic transformation \\
                       & Zoom blur          & Spatter          & Pixelation             \\
                       &                    &                  & JPEG compression       \\
        \bottomrule
    \end{tabular}
\end{table}
\subsection{VisDA-C}
VisDA-C\citep{peng2018visda} is a large-scale dataset to benchmark unsupervised domain adaptation methods. It
consists of three domains of 12 classes (object categories). The source (training) data consists renderings of 3D models from
various view-points and lighting conditions. The validation domain consists of real images cropped from MS-COCO. The
target (test) data also consists real images cropped Youtube bounding box datasets. All three datasets contain the
classes: aeroplane, bicycle, bus, car, horse, knife, motorbike, person, plant, skateboard, train, and truck.

\section{Detailed results of CIFAR datasets}\label{app:sec:cifar}
\begin{figure}
    \includegraphics[width=\textwidth]{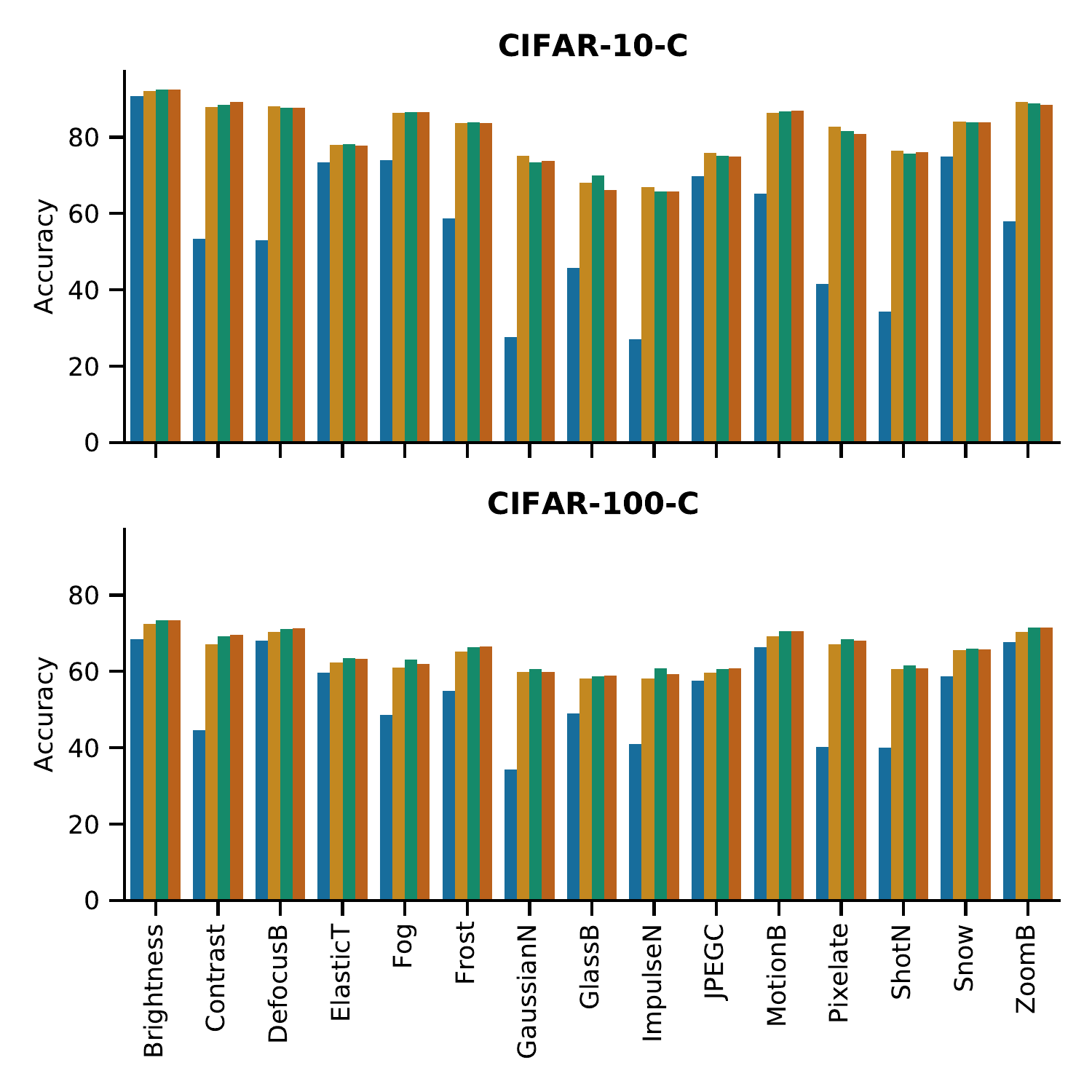}
    \caption{Performance comparisons for CIFAR-10-C and CIFAR-100-C datasets for \textcolor{color1}{unadapted},
        \textcolor{color2}{Tent}, \textcolor{color3}{Augmix}, \textcolor{color4}{Randaugment}  variants of our method. Our
proposed method, independent of the augmentation strategy used, gives a considerable improvement over the baseline. It
is also on par or slightly better than Tent for all corruption categories.}
    \label{app:fig:cifar}
\end{figure}

We show the detailed per corruption results of our CIFAR-C experiments in \cref{app:fig:cifar}. We find that our method
improves drastically when the baseline performance is low.

\section{Ablations}\label{app:sec:ablations}

\begin{figure}
    \includegraphics[width=\textwidth]{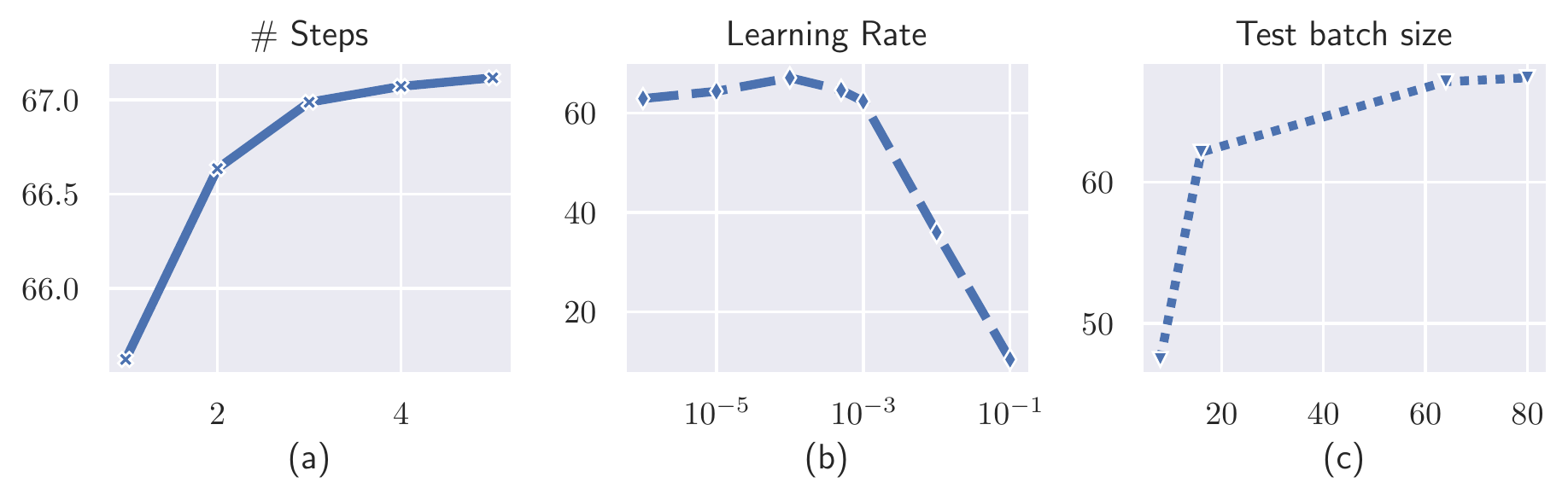}
\caption{Ablation experiments. With an unadapted performance of $44.1\%$, we show the importance of the factors. We see
an asymptotically improving performance with the number of SGD steps used. In (b), we see that our method is relatively
stable over a wide range of learning rates. Our method is dependent on batch size as in (c). The necessity of larger
batch size can pose constraints on the direct applicability of our method. This method }\label{app:fig:ablations}
\end{figure}

In this section, we provide ablations of three of the important hyperparameters of our algorithm, number of SGD steps,
learning rate, and testing batch size, here. We use the VisDA-C experimental setup and show our results for RandAugment
setup with default hyperparameters. The other hyperparameters remain the defaults set in the main paper, unless they are
being ablated on. The three ablations refer to \cref{app:fig:ablations}. For all the ablations in this section, we show
performance as $\%$ accuarcy on the test set.
\subsection{Number of SGD update steps}
The loss function in \cref{eqn:total} is minimized over several SGD steps, and we plot the evolution of performance with
number of steps. We see an asymptotic rise in the performance with number of steps, and we use $5$ steps for all other
steps. However this increases the run-time of our algorithm at test time, and thus we recommend a number dependent on
the running time requirements.
\subsection{Learning rate}
We change our learning rate on a logarithmic scale from $10^{-6}$ to $10^{-1}$, and we see that our method is reasonably
robust to the choice, with the best learning rate is around $10^{-4}$.
\subsection{Batch size}
We find that our method is stable for range of batch sizes, but does have a lower threshold around $16$. We hypothesize
this is due to training the normalization layers in the networks used. We were limited by the hardware available to run
larger batch sizes.

\subsection{Ablation of terms in Equation \ref{eqn:total}}
Our loss function in \cref{eqn:total} is a summation of two terms. Here we present the results obtained by considering
each of them terms. We see that in \cref{app:tab:lossabls}, the usage of \cref{eqn:consloss} leads to a performance
higher than that of Tent, and the combination of \cref{eqn:consloss,eqn:entropyloss} leads to better results, albeit at
a higher run-time requirement.

\begin{table}
    \centering
    \caption{Ablations of the loss terms in \cref{eqn:total}. }\label{app:tab:lossabls}
    \begin{tabular}{ccc}
        \toprule
        Unadapted & $\mathcal{L}_{cons}$ & $\mathcal{L}$ \\
        \midrule
        44.08     & 63.41                & 67.1          \\
        \bottomrule
    \end{tabular}
\end{table}

\subsection{Augmentation parameters}
\begin{figure}
    \centering
    \includegraphics[height=0.9\textheight]{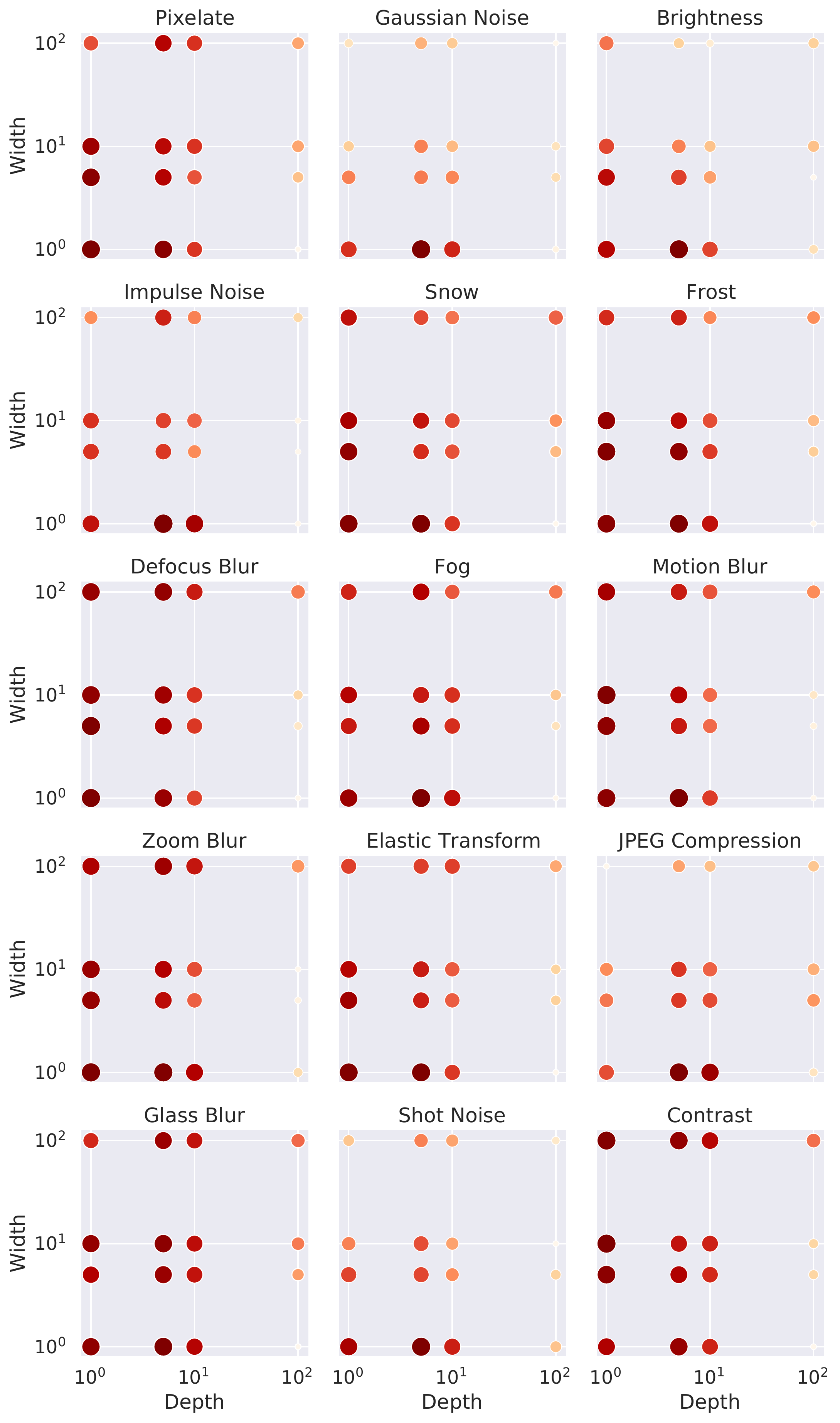}
    \caption{AugMix hyperparameters for CIFAR-10-C. We see that lesser augmentations result in better results.}
    \label{app:fig:augmix}
\end{figure}

\begin{figure}
    \centering
    \includegraphics[height=0.9\textheight]{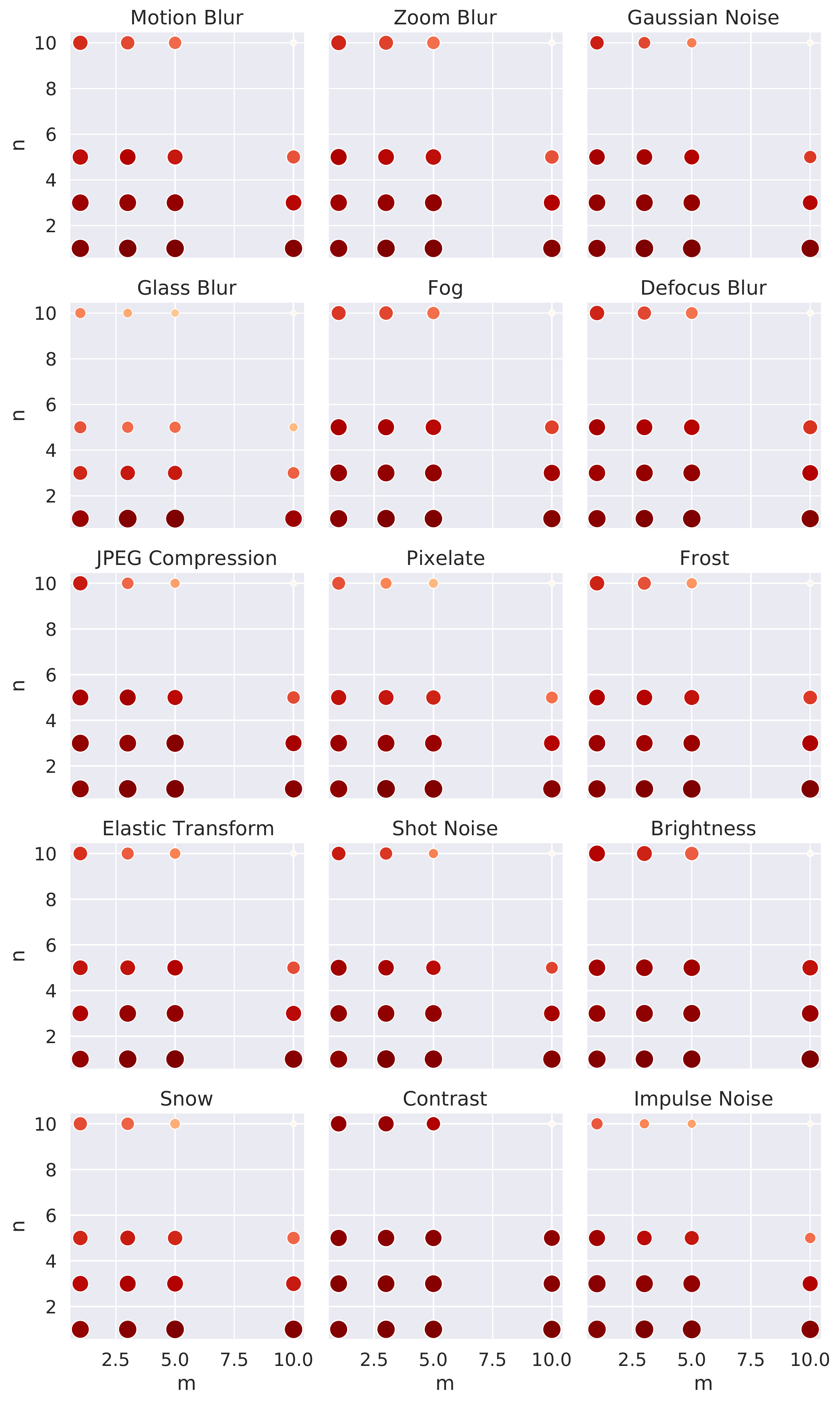}
    \caption{RandAugment hyperparameters for CIFAR-10-C. We see that lesser augmentations result in better results.}
    \label{app:fig:randaug}
\end{figure}

We vary the augmentation hyperparameters for both RandAugment, and AugMix in \cref{app:fig:augmix,app:fig:randaug}. For
this, we show results on CIFAR-10-C for each of the corruptions. We show results with only consistency loss
(\cref{eqn:consloss}), and width and brightness of each circle represents the overall final accuracy. We find that for
nearly all the corruptions, lower amount of augmentation results in better performance compared to ones with higher
augmentation. We hypothesize this is because lower intensity augmentations result in samples closer to the input sample,
and thus results in samples from the image manifold.


\end{document}